\documentclass[]{interact}
\usepackage{wrapfig}
\usepackage{epigraph}
\usepackage{epstopdf}
\usepackage[caption=false]{subfig}

\theoremstyle{plain}

\theoremstyle{definition}

\theoremstyle{remark}

\usepackage[natbibapa,nodoi]{apacite}
\usepackage{natbib}
\setcitestyle{authoryear,open={(},close={)},citesep={;},aysep={,},yysep={;},maxcitenames=1}

\usepackage{float}
\usepackage[dvipsnames]{xcolor} 
\usepackage{amssymb}
\usepackage{graphicx}
\usepackage{balance}
\usepackage{multirow}
\usepackage{caption}
\usepackage[tight,english]{minitoc}

\usepackage{epsfig,amsfonts,amsmath, cases,graphics, latexsym, times, algorithmic, algorithm, bbm, soul,wrapfig,comment}

\usepackage{smartdiagram}
\usepackage[draft]{todonotes} 
\usepackage{verbatim}
\usepackage{listings}
\usepackage{color}
\usepackage{adjustbox} 
\usepackage{listings}
\usepackage{rotating} 

\usepackage[final]{pdfpages} 

\usepackage{textcomp}
\newcommand{\greencheckmark}{\textcolor{green}{\checkmark}}
\newcommand{\redxmark}{\textcolor{red}{\texttimes}}

\usepackage[T1]{fontenc} 
\definecolor{gray}{rgb}{0.5,0.5,0.5}
\definecolor{mauve}{rgb}{0.58,0,0.82}
\definecolor{gray}{rgb}{0.4,0.4,0.4}
\definecolor{darkblue}{rgb}{0.0,0.0,0.6}
\definecolor{lightblue}{rgb}{0.0,0.0,0.9}
\definecolor{cyan}{rgb}{0.0,0.6,0.6}
\definecolor{darkred}{rgb}{0.6,0.0,0.0}
\definecolor{darkblue}{rgb}{0.0,0.0,0.6}
\definecolor{dkgreen}{rgb}{0.0, 0.29, 0.33}

\lstdefinelanguage{turtle}
{
    columns=fullflexible,
    keywordstyle=\color{red},
    morekeywords={PREFIX,SELECT,DISTINCT,UNION,FILTER,ORDER,BY,REGEX,STR,STRSTARTS,GRAPH,isBlank},
    morecomment=[l]{\#},
    tabsize=4,
    frame=lines,
    numbers=left,
    numberfirstline=true,
    xleftmargin=2.5em,
    framexleftmargin=2.8em,
    stepnumber=1,    
    firstnumber=1,
    alsoletter={-?}, 
    morecomment=[s][\color{blue}]{<}{>},
    commentstyle=\color{green!40!black},
    basicstyle=\scriptsize\ttfamily\color{black},
    morestring=[b][\color{black}]\",
    backgroundcolor=\color{background},    
    showstringspaces=false
}

\lstdefinelanguage{XML_LNG}
{
  morestring=[s][\color{darkblue}]{"}{"},
  morestring=[s][\color{black}]{>}{<},
  morecomment=[s]{<?}{?>},
  morecomment=[s][\color{dkgreen}]{<!--}{-->},
  stringstyle=\color{mauve},
  identifierstyle=\color{dkgreen},
  keywordstyle=\color{darkred},
  morekeywords={bn,n,t,u,v, semanticAnnotationType, deduce, suggest, suggest_comment, deduceUri, suggestUri}
}

\lstdefinelanguage{RULE_LNG}
{
  stringstyle=\color{black},
  identifierstyle=\color{black},
  keywordstyle=\bfseries\color{darkblue},
  morekeywords={lessThan, greaterThan}
}

\lstdefinelanguage{SPARQL_LNG}
{
  stringstyle=\color{black},
  identifierstyle=\color{black},
  keywordstyle=\bfseries\color{darkblue},
  morekeywords={PREFIX,SELECT,DISTINCT,WHERE,OPTIONAL,FILTER,LANGMATCHES,LANG}
}

\lstdefinelanguage{JAVA_LNG}
{
  morecomment=[s][\color{dkgreen}]{/**}{*/},
  stringstyle=\color{black},
  identifierstyle=\color{black},
  keywordstyle=\bfseries\color{darkred},
  morekeywords={public,new,return,true,this}
}

\usepackage{url}

\renewcommand\bibliographytypesize{\fontsize{10}{12}\selectfont}

\begin{document}

\articletype{ARTICLE TEMPLATE}

\title{IoT-Based Preventive Mental Health Using Knowledge Graphs and Standards for Better Well-Being}
%



\author{
\name{Amelie Gyrard\textsuperscript{a,c,*}\thanks{CONTACT A.~N. Author. Email: amelie.gyrard@trialog.com}, Seyedali Mohammadi\textsuperscript{b}, Manas Gaur\textsuperscript{b} and Antonio Kung\textsuperscript{a} }
\affil{\textsuperscript{a}Trialog, Paris, France; \newline
\textsuperscript{b}University of Maryland, Baltimore County (UMBC), USA; 
\newline\textsuperscript{c}Machine-to-Machine Measurement (M3), Paris, France}
}

%
\maketitle

\begin{abstract}
Sustainable Development Goals (SDGs) give the UN a road map for development with Agenda 2030 as a target.
SDG3 "Good Health and Well-Being" ensures healthy lives and promotes well-being for all ages. Digital technologies can support SDG3.
Burnout and even depression could be reduced by encouraging better preventive health. Due to the lack of patient knowledge and focus to take care of their health, it is necessary to help patients before it is too late. New trends such as positive psychology and mindfulness are highly encouraged in the USA.
\textbf{Digital Twins (DTs)} can help with the continuous monitoring of emotion using physiological signals (e.g., collected via wearables). DTs facilitate monitoring and provide constant health insight to improve quality of life and well-being with better personalization.
Healthcare DTs challenges are standardizing data formats, communication protocols, and data exchange mechanisms. As an example, ISO has the ISO/IEC JTC 1/SC 41 Internet of Things (IoT) and DTs Working Group, with standards such as "ISO/IEC 21823-3:2021 IoT - Interoperability for IoT Systems - Part 3 Semantic interoperability", "ISO/IEC CD 30178 - IoT - Data format, value and coding". To achieve those data integration and knowledge challenges, we designed the \textbf{Mental Health Knowledge Graph} (ontology and dataset) to boost mental health. As an example, explicit knowledge is described such as chocolate contains magnesium which is recommended for depression. The Knowledge Graph (KG) acquires knowledge from ontology-based mental health projects classified within the LOV4IoT ontology catalog (Emotion, Depression, and Mental Health). Furthermore, the KG is mapped to \textbf{standards} (e.g., W3C Semantic Web languages such as RDF, RDFS, OWL, SPARQL to develop and query ontologies and datasets) when possible. Standards from ETSI SmartM2M can be used such as SAREF4EHAW (SAREF for eHealth Ageing Well domain) to represent medical devices and sensors, but also ITU/WHO, ISO, W3C, NIST, and IEEE standards relevant to mental health can be considered.
\end{abstract}

\epigraph{"An ounce of prevention is worth a pound of cure."}{ Benjamin Franklin} 

\epigraph{"How to gain, how to keep, how to recover happiness is in fact for most men at all times the secret motive of all they do, and of all they are willing to endure."}{ William James} 


\section{Introduction}
\label{section:Introduction}




More than one-fifth of adults in the United States have dealt with mental health issues, according to the National Institute of Mental Health\footnote{\url{https://www.nimh.nih.gov/health/statistics/mental-illness}}. This situation has led to the government setting aside \$280 billion to improve the availability and quality of mental health services\footnote{\url{https://www.whitehouse.gov/cea/written-materials/2022/05/31/reducing-the-economic-burden-of-unmet-mental-health-needs/}}. 
Mental health can increase productivity and efficiency, improve staff morale, and reduce absenteeism \citep{albraikan2019inharmony}.
There are numerous reviews on mental health using Wearable sensors and Artificial Intelligence Techniques \citep{gedam2021review}.
\textbf{Sustainable Development Goals (SDGs)}\footnote{\url{https://sdgs.un.org/goals}} give the UN a road map for development with Agenda 2030 as a target.
SDG3 "Good Health and Well-Being" ensures healthy lives and promotes well-being for all ages. Digital technologies can support SDG3.
We review hereafter definitions relevant to mental health, the need for Digital Twin for Mental Health, and the benefit of a Knowledge Graph (KG).
\textbf{Mental health}\footnote{\url{https://www.who.int/health-topics/mental-health#tab=tab_1}} is a state of mental well-being that enables people to cope with the stresses of life, realize their abilities, learn well, and work well, and contribute to their community. It has intrinsic and instrumental value and is integral to our well-being.
\textbf{IEEE 7010} defines \textbf{well-being} as "the continuous and sustainable physical, mental, and social flourishing of individuals, communities, and populations where their economic needs are cared for within a thriving ecological environment."
\textbf{Stress} vs. \textbf{Anxiety:} People under stress\footnote{\url{https://www.apa.org/topics/stress/anxiety-difference}} experience mental and physical symptoms, such as irritability, anger, fatigue, muscle pain, digestive troubles, and difficulty sleeping. Anxiety, on the other hand, is defined by persistent, excessive worries that don't go away even in the absence of a stressor\footnote{\url{https://www.apa.org/topics/stress/anxiety-difference}}.
\textbf{Burnout} and even \textbf{depression}\footnote{\url{https://www.who.int/news-room/fact-sheets/detail/depression}} could be reduced by encouraging better preventive health. Lack of patient knowledge and focus to take care of their health before it is too late. New trends such as positive psychology \citep{seligman2008positive} and mindfulness (MBSR) \citep{kabat2003mindfulness} are highly encouraged in the USA. 
Depression is considered the main mental health crisis by the \textbf{World Health Organization (WHO)} \citep{mullick2022iot}. Mental health includes emotional, psychological, and social well-being changes.
\textbf{DSM-V (Diagnostic and Statistical Manual of Mental Disorders)}\footnote{\url{https://www.psychiatry.org/psychiatrists/practice/dsm}} references more than 70 mental disorders that complement the International Classification of Diseases (ICD). DSM-V helps clinicians and researchers define and classify mental disorders, which can improve diagnoses, treatment, and research. DSM-V provides a form with checklists of symptoms for better diagnosis. Burnout does NOT appear in DSM and ICD. 
 \textbf{Mental illness} is a type of health condition that changes a person’s mind, emotions, or behavior and has been shown to impact an individual’s physical health \citep{su2020deep}.
\textbf{Depressive disorders}, or unipolar depression's symptoms are low mood, loss of interest in day-to-day activities, significant weight changes, reduction of mobility, constant fatigue, difficulty concentrating, and feelings of worthlessness that can be diagnosed, and severity, using the Patient Health Questionnaire-9 (PHQ-9) \citep{gutierrez2021internet}.

\textbf{The necessity of AI-enabled IoT Digital Twin using knowledge graph for achieving SDG-3:} An IoT Digital Twin designed for proactive mental health care corresponds with four of the United Nations' Sustainable Development Goals. First is good health and well-being (SDG-3), which aims to ensure healthy lives and promote well-being for all age groups. We provide a pragmatic approach to developing IoT Digital Twin using domain-specific standards and knowledge graphs, which would advance mental health care and well-being. Disparities in mental health care access are prevalent, varying among regions, socioeconomic statuses, and genders. By prioritizing preventive strategies using domain-specific standards (e.g., questionnaires, guidelines) and leveraging AI-enabled IoT technology, this endeavor can potentially mitigate these access gaps, aligning with initiatives like the NIH's AIM-AHEAD Initiative.

\textbf{Why do we need a Digital Twin for Mental Health?}
The \textbf{StandICT landscape of Digital Twins (DT)}\footnote{\url{https://www.standict.eu/landscape-analysis-report/landscape-digital-twin}} reminds that "Digital Twin" was first introduced by Professor Michael Grieves from the University of Michigan in 2002. According to \textbf{ISO/IEC 30173 Digital twin – concepts and terminology}\footnote{\url{https://www.iso.org/standard/81442.html}}, DTs are defined as a "digital representation of a target entity with data connections that enable convergence between the physical and digital states at an appropriate rate of synchronization."
Notable implementation of DTs include Siemens Healthineers \citep{erol2020digital}, IBM Maximo
Application Suite\footnote{\url{https://www.ibm.com/products/maximo}} and Philips HeartModel\footnote{\url{https://www.philips.com/a-w/about/news/archive/blogs/innovation-matters/20181112-how-a-virtual-heart-could-save-your-real-one.html}}. DT can help with emotion monitoring using physiological signals (e.g., collected via wearables).
Healthcare DTs facilitate monitoring, understanding, and optimization of human functioning and monitor health to improve quality of life and well-being \citep{laamarti2020iso}, \citep{ferdousi2022digital},  \citep{albraikan2019inharmony}, \citep{anand2024hygieia} with better personalization \citep{bagaria2020health} by using digital technologies such as IoT, AI, etc. Healthcare DT challenges highlight the need for standardizing data format, communication protocols, and data-exchange mechanisms \citep{turab2023comprehensive}.

\textbf{Why do we need Knowledge Graphs for Mental Health?} \textbf{The European Human Brain Project (150 000 000 Euros, 161 partners), more precisely, the EBRAINS KG}\footnote{\url{https://search.kg.ebrains.eu/?facet_type[0]=Dataset&q=cortisol}}, is a synergy project between neuroscience, computing, informatics, and brain-inspired technologies, encourages open-science through a web-based system to share tools, etc. However, when looking for keywords such as cortisol (the stress hormone), nothing can be found.
KGs \citep{ShethKGKnoesis2019} can use ontologies to structure data. An ontology provides a shared common understanding of a domain \citep{gruber1995toward}. The use of ontologies is already recognized in the biomedical domain with BioPortal ontology catalog\footnote{\url{https://bioportal.bioontology.org/}}.
\textbf{Ontology Development 101 methodology} (designed by the creators of the Standford Protégé Ontology Editor tool) encourages reusing domain knowledge by reusing ontologies in Step 2 "Consider reusing ontologies."

 In this paper, we describe about semantic interoperability applied to mental health, which is a follow-up of our health IoT semantic interoperability past work \citep{gyrardhappinesschase2019} \citep{gyrardkhealthreasonerbookchapter3}.

\begin{wrapfigure}{r}{0.62\textwidth}
 \footnotesize
 \centering
 \includegraphics[width=0.57\textwidth]{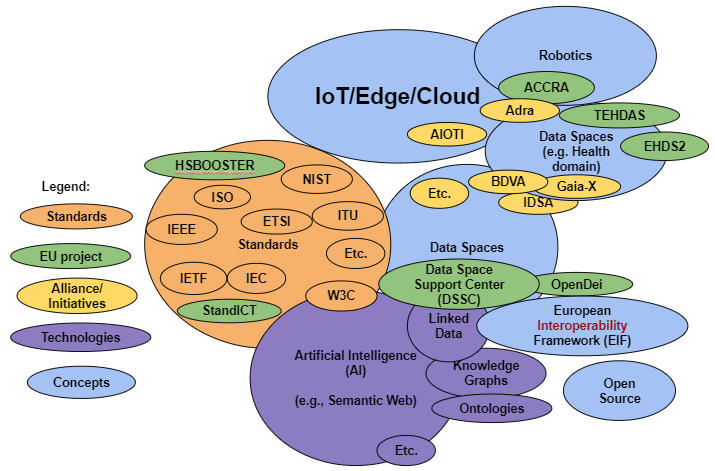}
 \caption{Standards, Semantic Web and Data Spaces applied to Health}
 \label{figure:DataspaceRobotics} 
 \end{wrapfigure}

\textbf{Toward Converging Standards and Semantic Web Technologies applied to Mental Health:} We illustrate the vision of our aligning technologies on Semantic Web and Data spaces and Standards such as ISO, IEC, and W3C as depicted in Figure \ref{figure:DataspaceRobotics}. W3C Standards are more adapted to Semantic Web technologies such as W3C RDF, W3C RDFS, W3C OWL, W3C SPARQL, etc. Semantic Web technologies include knowledge Graphs with ontologies and Linked Data. There are also ISO standards about ontologies such as ISO/IEC 21838-1:2021 Information technology — Top-level ontologies (TLO) — Part 1: Requirements, ISO/IEC 21838-2:2021
Information technology — Top-level ontologies (TLO) — Part 2: Basic Formal Ontology (BFO), etc. Data Space initiatives such as BDVA, Gaia-X, etc. are considered. Some energy data spaces are designed by projects such as EDHS2, and TEHDAS. 
The list is not exhaustive.
"Charting past, present, and future research in the semantic web and interoperability" highlights the importance of semantic interoperability \citep{rejeb2022charting} and on our past work: Table 5 "Top 20 most productive authors" and Figure 4 "Co-authorship network."
The recommendation is that interoperability should remain a primary research focus and closer collaborations with industry, governments, and standards organizations to develop more effective models.

Our vision has been shared with the EUCloudEdgeIoT Concentration and Consultation Meeting on Computing Continuum: Uniting the European ICT community for a digital future in May 2023\footnote{\url{https://eucloudedgeiot.eu/concentration-and-consultation-meeting-on-computing-continuum-uniting-the-european-ict-community-for-a-digital-future/}}.

The remainder of this chapter is organized as follows: related work limitations in Section~\ref{section:RelatedWork},  standards for mental health in Section~\ref{section:standards}, the mental health ontology catalog in Section~\ref{section:ontologycatalog} (which includes mapping to standards ontologies), the project use cases in Section~\ref{section:UseCases}, and conclusion in Section~\ref{section:ConclusionFutureWork}.
\section{Related Work: IoT, Digital Twin, and AI for Mental Health, Stress and Well-Being}
\label{section:RelatedWork}
This section reviews IoT-based Mental Health Monitoring Applications, IoT-based Stress Detection, Digital Twin for Mental Health, and AI-based Mental Health.

\begin{table}[!htbp]
\resizebox{\textwidth}{!}{%
\begin{tabular}{ | l | l | l | l | l |}
    \hline
     \textbf{Authors} &  \textbf{Year} & \textbf{Research Problem } & \textbf{Sensor or } & \textbf{Reasoning}\\
     
       & & \textbf{Addressed \& Project}&  \textbf{Measurement Type} & \\

         \hline
        \hline 
   \citeauthor{garcia2018mental} & 2018 &\textbf{Mental Health Monitoring} & \greencheckmark Heart rate, GSR, & Survey paper \\
     & &  {Systems (NHMS)} Survey & body or skin temperature  &\greencheckmark ML algorithm \\

     \hline 
     \citeauthor{kim2017unobtrusive} & 2017 & \textbf{Depression} Severity & \greencheckmark Infrared motion sensor & \greencheckmark Bayesian Network\\
       & & Elderly People, AAL &  & Decision Tree, SVM, ANN  \\
    
    \hline 
    \citeauthor{zhou2015tackling} & 2015 & Monitoring \textbf{mental health}  & \greencheckmark Heart rate, pupil variation, & \greencheckmark ML (Logistic  \\
   &  & states & head movement, eye blink, facial expression & regression, SVM)\\
        
     
 \hline 
    \hline 
   \citeauthor{garcia2016automatic}   & 2016 & \textbf{Stress} & \greencheckmark Accelerometer data (from smartphone)  & \greencheckmark Naive Bayes, Decision Tree \\
      
   \hline 
     \citeauthor{yoon2016flexible} & 2016 & New \textbf{stress} monitoring patch & \greencheckmark Skin conductance, pulse wave, skin temperature& \redxmark - \\
 
		  \hline 
      \citeauthor{lu2012stresssense} & 2012 & \textbf{Stress}Sense & \greencheckmark Voice data (smartphone)  & \greencheckmark GMMs \\
 
     \hline 
      \citeauthor{chang2011s} & 2011 & AMMON: \textbf{Stress} detector & \greencheckmark Voice data & \greencheckmark SVM \\
    

\hline
    \hline 
 \citeauthor{erol2020digital} & 2020  & \textbf{DT} for patients and medical devices  & \redxmark No & \redxmark No  \\

    \hline 
 \citeauthor{liu2019novel} & 2019  & CloudDTH - \textbf{DT} healthcare & \greencheckmark Yes Wearable medical devices & \redxmark No \\

    \hline 
 \citeauthor{albraikan2019inharmony} & 2019  & inHarmony - Emotional Well-being \textbf{DT}  & \greencheckmark Yes Empatica E4 & \redxmark No\\

 \hline	
\end{tabular}}
\caption{Digital Twin for Mental Health, Depression, Stress, and  Well-Being: sensors, reasoning, and applications.}
\label{table:DigitalTwinHealthcareSOTA}
\end{table}

\textbf{IoT-based Mental health and depression monitoring applications} are summarized in Table \ref{table:DigitalTwinHealthcareSOTA}. \textbf{23 Mental Health Monitoring Systems (NHMS) \cite{garcia2018mental}} using sensor data provided by IoT devices and analyzed with Machine Learning are classified according to the study type (e.g., bipolar disorder detection, migraine forecasting, depression detection, anxiety detection, stress detection, social phobia association, bipolar disorder association, anxiety association, depression association, and epilepsy). Various devices are employed (e.g., eye sensor, heart rate variability, electrodermal activity, wrist accelerometer, galvanic skin response, skin conductance, spo2, photoplethysmogram, accelerometer, gyroscope, pressure sensitive, video cameras, VR headset, pupil-corneal reflection and head tracker, SMS, calls, screen, GPS, location, touchscreens, audio, contacts, videos, sound, head-mounted display, questionnaires). 
\textbf{Conclusion: We focus on IoT devices relevant to mental health disorders such as stress, anxiety, and depression, and do not consider others such as bipolar, migraine, or phobia. The authors do not mention at all the term "ontology" or knowledge graph." }

\textbf{24  IoT-based mental health care applications \cite{gutierrez2021internet} (bipolar disorders, depression, schizophrenia, and stress-related disorders)}; from 2010 to 2020 are designed for: 
1) data acquisition, 
2) self-organization, 
3) service level agreement, and 
4) identity management during mental health interventions. 
IoT devices considered are proximity sensors, ambient light, accelerometers, gyroscopes, magnetometers, ambient sound, barometers, temperature, and humidity. Diagnostic and Statistical Manual of Mental Disorders 5 (DSM 5) categorizes the mental health disorder literature.
\textbf{HealthyOffice smartphone app \cite{zenonos2016healthyoffice}} focuses on eight mood state recognition (Excited, Happy, Calm, Tired, Bored, Sad, Stressed, Angr) in work environments. 

\textbf{Conclusion: We focus on mental health disorders such as stress, and depression, and do not consider other disorders such as bipolar, or schizophrenia. Gutierrez et al. \cite{gutierrez2021internet} only cite the mental health ontology from Hazdic et al. \cite{hadzic2008towards}, and do not mention at all the KG term, so key research on depression ontologies and KGs are missing. For this reason, we built the ontology catalog as introduced in Section \ref{section:MentalHealthOntologyCatalog} and Table \ref{table:depressionOntologySOTA}.} 

IoT-based Stress Detection is summarized in Table \ref{table:DigitalTwinHealthcareSOTA}.
Stress is a reaction from the human body in response to a challenging event or a demanding condition \cite{gedam2021review}; which can be detected using wearable sensors (e.g., Electrocardiogram, Electroencephalography, and Photoplethysmography) and applied machine learning. Activities such as driving, studying, and working are taken into consideration. 
\textbf{Wearable healthcare-monitoring systems for pain and stress detection} \cite{chen2021pain}) are using wearable sensors to measure physiological signals such as heart, brain, muscle, electrodermal, respiratory, blood volume pulse, and skin temperature.
A system to detect, diagnose and manage mental health emergencies \cite{mullick2022iot}; is based on four physiological parameters: heart rate, Spo2, body temperature (LM35 sensor), pressure (BMP180 sensor).

Digital Twin (DT) for Mental Health and  Well-Being is summarized in Table \ref{table:DigitalTwinHealthcareSOTA}.
\textbf{Digital Twin Coaching for Physical Activities's survey \citep{gamez2020digital}} explore papers from Scopus, Web of Science, IEEE Xplore and ACM Digital Library the last ten years (2010–2020).  The papers have been classified into: 1) sports, 2)
wellbeing, and 3) rehabilitation. The following information is extracted: 
1) ML algorithms being used 
2)	Type of application being researched,
3)	Sensors and actuators devices used,
4)	Performance of the ML algorithm,
5)	Usability feedback of users about the system.
\textbf{Digital Twin in healthcare \citep{turab2023comprehensive}} highlights the need for standardizing data format, communication protocols, and data-exchange mechanisms. Turab et al. highlight that ISO has an interest in Digital Twin.
Indeed, we are involved in ISO SC 41 IoT and Digital Twin.
\textbf{Well-being digital twin (WDT) \citep{ferdousi2022digital}} challenges for predictive well-being applications are technical issues of handling heterogeneous data and standards, data bias, level of autonomy, trust in intelligence, data visualization issues, and consent of humans.
Drawbacks of WDT are:
1) Inadequate or missing data,
2) Ethical overheads,
3) Trust in AI, 
and 4) Necessity of domain knowledge.
Well-being digital twin (WDT) areas are:
1) Collecting and managing vast healthcare data,
2) Meaningful data visualization,
3) Facilitating predictive healthcare,
4) Improving healthcare quality of experience (QoE),
5) Personalized healthcare system (e.g., digital coaching, elderly healthcare (e.g. device in \cite{ponmalar2024iomt}), immune system care), 
and 6) Understanding clinical pathways.
An ISO/IEEE 11073 standardized digital twin framework architecture for health and well-being to analyze data from personal health devices is compliant or not with X73 standards \citep{laamarti2020iso}.
\textbf{inHarmony \citep{albraikan2019inharmony}} is an Emotional Well-being DT for workplace which includes emotion detection, emotional biofeedback, and emotion-aware recommender systems. A usability study was carried out with 35 practitioners wearing the Empatica E4 sensor to acquire physiological signals.
\textbf{CloudDTH \citep{liu2019novel}} is a DT Healthcare (DHT) framework for real-time monitoring, diagnosing, and predicting of elderly's health, based on cloud, and wearable medical devices technologies. A Big Data-based Precision Medicine Cloud Platform is designed to display human physiological health data for data management, data analysis, etc.
DTs of patients and DTs of medical devices are considered \citep{erol2020digital}.
DT \textbf{Siemens Healthineers} \citep{erol2020digital} for radiology department in Mater Private Hospitals from Ireland  to reduce patient's waiting time.
Siemens Healthineers is also applied to cardiologists in a research project at Heidelberg University for patients with chronic congestive heart failure.
\textbf{Philips HeartModel}\footnote{\url{https://www.philips.com/a-w/about/news/archive/blogs/innovation-matters/20181112-how-a-virtual-heart-could-save-your-real-one.html}} is a personalized Digital Twin of the heart to help surgeons with real-time 3D. Its HeartNavigator tool combines the Computed Tomography (CT) heart anatomy obtained before the surgical procedure and live during the surgery. 
\textbf{Sooma} is a startup that simulates the electrical
signals of the brain to treat depression, other neurological and
psychiatric disorders.

\textbf{Conclusion:} We did not find a digital twin for mental health. inHarmony is addressing well-being.

\textbf{AI-based Mental Health:} "\textbf{Positive AI" \citep{van2023positive}}, means that AI systems actively and intentionally support human well-being based on fields such as positive psychology, human-centered design, and computing. The authors categorized twelve well-being AI challenges into 1) knowledge (how to conceptualize, operationalize, optimize for, and design), and 2) motivation (misaligned incentives, PR \& monetary risks, and lack of access, preventing).
To identify mental health states, machine learning algorithms such as support vector machines, decision trees, naïve Bayes classifier, K-nearest neighbor classifier, and logistic regression are used \citep{srividya2018behavioral}.

 \textbf{Related work shortcomings: }
 Those works do not use health standards that we are addressing in Section \ref{section:standards}.
 Furthermore, in this chapter, we will focus on KGs (e.g., ontologies) as AI solutions as explained in Section \ref{section:ontologycatalog}.

\section{Standards for Mental Health and Well-Being: Artificial Intelligence (Semantic Web, Ontologies)}
\label{section:standards}


\begin{wrapfigure}{r}{.65\textwidth}
\footnotesize
\centering
\includegraphics[width=0.63\textwidth]{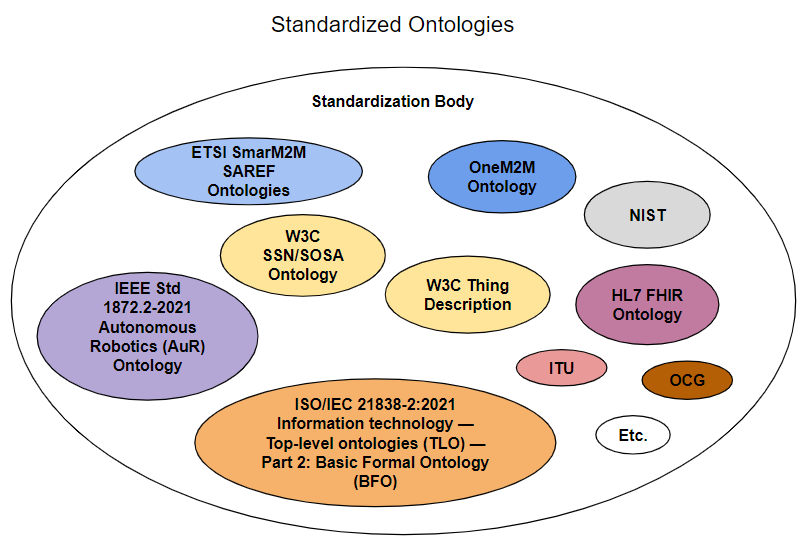}
\caption{Standardized Ontologies}
\label{StandardizedOntologies} 
\end{wrapfigure}
We investigate Standards Development Organizations (SDOs) for (mental) health covering AI (Semantic Web, Ontologies) such as:
ETSI SmartM2M EHealth/Ageing-Well Ontology in Section \ref{section:ETSI_SMARTM2M_SAREF_Health},
ITU/WHO Focus Group on Artificial Intelligence for Health in Section \ref{section:ITU_Health},
ISO Health Standards in Section
\ref{section:ISO_Health},
W3C Health Standards in Section
\ref{section:W3C_Health},
NIST Health Standards in Section \ref{section:NIST_Health}, and
IEEE Health Standards in Section
\ref{section:IEEE_Health}.
Standards Development Organizations SDOs define health ontologies or IoT ontologies describing sensors to measure the physiological signals of patients as depicted in Figure \ref{StandardizedOntologies}.


\subsection{ETSI SmartM2M SAREF for EHealth/Ageing-Well Ontology}
\label{section:ETSI_SMARTM2M_SAREF_Health}
\textbf{ETSI Smart M2M SAREF4EHAW ontology} \citep{SAREF4EHAW2020} reviews standards such as IEEE, ETSI, SNOMED, OneM2M), Alliances (AIOTI), IoT Platforms, and European projects and initiatives, etc. 

The use cases are classified into some of those categories:
1) Daily Activity Monitoring,
2) Cognitive simulation for mental decline prevention,
3) Prevention of social isolation,
4) Comfort and safety at home.
SAREF4EHAW investigated the following ontologies:
1) WSNs/measurement ontologies: OGC (Open Geospatial Consortium) Observations \& Measurements (O\&M), Sensor Model Language (SensorML), Semantic Sensor Web (SWE): W3C \& OGC SOSA (Sensing, Observation, Sampling, and Actuation) and W3C SSN (Semantic Sensor Network).
NASA QUDT (Quantities, Units, Dimensions, and Types).
2) eHealth/Ageing-well domain main ontologies:
ISO/IEEE 11073 Personal Health Device (PHD) standards,
ETSI SmartBAN Reference Data Model and associated modular ontologies,
FHIR RDF (Resource Description Framework),
FIESTA-IoT Ontology to support the federation of testbeds,
Bluetooth® LE (Low Energy) profiles for medical devices proposed by Zontinua, MIMU-Wear (Magnetic and Inertial Measurement Units) ontology, and Active and Healthy Ageing (AHA) platform wearables' device ontology.
The document explains that SAREF has been mapped with oneM2M base ontology in 2017.
SAREF4EHAW \citep{SAREF4EHAW2020} selected 43 ontological requirements, and 59 service-level assumptions of the eHealth/Ageing-well domain (use cases included). For instance, as a requirement, the ontology describes ECG (\citep{muthalagu2023pattern}) device concepts.
There is also SAREF4WEAR \footnote{\url{https://saref.etsi.org/saref4wear/}} an extension for wearables.


\textbf{Conclusion: SAREF4EHAW demonstrates the need for ontologies. We did not find standards more specific to mental health in ETSI.}


\subsection{ITU/WHO Focus Group on Artificial Intelligence for Health}
\label{section:ITU_Health}
\textbf{ITU/WHO Focus Group on Artificial Intelligence for Health (FG-AI4H)}\footnote{\url{https://www.itu.int/go/fgai4h}}, established in 2018, free and open, partners with the World Health Organization (WHO) to standardize AI-based assessment framework and evaluation for health, diagnosis, triage or treatment decisions. It provides an AI for health online platform and complementary tools for benchmarking of data. FG-AI4H comprises experts: machine learning/AI researchers, healthcare practitioners and researchers, regulators, representatives of health ministries and ministries of telecommunication, international organizations, and individuals from complementary fields. The FG highlights the need for explainability and interpretability of AI tools.
\textbf{ITU-T/WHO FG-AI4H use cases:}
24 AI4H use cases are introduced, including traditional medicine, psychiatry, neurological disorders, falls among the elderly, etc.
\textbf{DEL.10.23 TG-AI for Traditional Medicine} reviews the existing AI solutions for Traditional Medicine. AI in traditional medicine will be relevant to support healthcare practitioners in recommending integrative medical practices. Traditional Medicine treatments: 1) Ontological and Natural language processing (NLP) are encouraged to extract useful information, 2) Analyzing and integrating data, and 3) Decision support, 4) Predictive analytics, 5) Patient monitoring and feedback, 6) virtual health assistants, and 7) Research and knowledge discovery.

\textbf{Conclusion: We find a standard on AI Traditional Medicine, but we did not find standards for mental health in this SDO.}

\subsection{ISO Health Standards}
\label{section:ISO_Health}

The \textbf{CEN/ISO EN13606} is a European norm approved as an ISO standard to achieve EHR semantic interoperability.
\textbf{ISO 13606-5:2010 Health Informatics - Electronic Health Record communication standards}\footnote{\url{http://www.en13606.org/information.html}} defines an architecture for exchanging Electronic Health Record (EHR) describing patient's health status and ease communication between EHR systems (e.g., clinicians applications, decision support systems).
ISO 13606 seems an open standard.
Within the \textbf{ISO SC41 IoT and Digital Twin}, there is an ongoing standard entitled "\textbf{IoT/IEC 30197 IoT for Stress Management, Good health \& Well-being}".
\textbf{ISO SC42 AI - "AI-enabled Health Informatics"}
is a joint working group that will provide a landscape survey and a set of recommendations for future work on the impact of ISO/TC 215 standards. The group comprises experts from ISO/IEC/JTC 1/SC 42, IEEE, and the ITU/WHO AI4Health focus group.
\textbf{ISO/IEEE 11073 medical device communication standard - Used by Personal Connected Health Alliance (PCHAlliance) - X73 standards} facilitates health data exchange while providing plug-and-play real-time interoperability by being compliant with the X73 communication model. However, the X73 standards do not address security or users’ privacy.
This ISO/IEEE 11073 standard is used by \cite{laamarti2020iso} for a digital twin framework architecture for health and well-being in smart cities.
NIST researchers are collaborating with medical device experts to develop standards for medical device communications to enhance semantic interoperability. XML schema and tool based on the ISO/IEEE 11073 medical device communication standard is developed by Garguilo et al. \citep{garguilo2007moving}.
Personal Connected Health Alliance (PCHAlliance) products use the ISO/IEEE 11073 Personal Health Data (PHD) Standards.

\textbf{Conclusion: ISO 215 Health Informatics must be explored further to dig any documents related to mental health. CEN/ISO EN13606 focuses on EHR semantic interoperability. ISO SC42 AI joint working group “AI-enabled Health Informatics” comprises experts from ISO/IEC/JTC 1/SC 42, IEEE, and the ITU/WHO AI4Health focus group.}

\subsection{W3C Semantic Web Health Care and Life Sciences Community Group (HCLS CG)}
\label{section:W3C_Health}

\textbf{W3C Semantic Web (SW) Health Care and Life Sciences Community Group (HCLS CG)}\footnote{\url{https://www.w3.org/community/hclscg/}} encourages the use of SW technologies across health care, life sciences, clinical research and translational medicine as they need interoperability of information from many disciplines. 
HCLS CG use SW technologies to design use cases which have a clinical, research of business values. The CG develops liaisons with organizations in healthcare, life sciences, and clinical research, including organizations that are working on standards.

\textbf{Conclusion: W3C HCLS CG highlights the need of using Semantic Web. The final report is more focused on drug-drug interaction. We did not find standards specific to mental health. }


\subsection{NIST Health Standards}
\label{section:NIST_Health}


NIST researchers are collaborating with medical device experts to develop standards for medical device communications to enhance semantic interoperability. XML schema and tool based on the ISO/IEEE 11073 medical device communication standard \citep{garguilo2007moving}.

\textbf{Conclusion: NIST has an interest in semantic interoperability.}

\subsection{IEEE Health Standards}
\label{section:IEEE_Health}

IEEE digital health standards can help save lives and improve people's quality of life.
Standards aim to share information across end-to-end infrastructure, particularly: 1) interoperability: information exchange among organizations,
2) cost savings/efficiency (e.g., for hospitals), etc.
We selected four standards:
1) IEEE 7010’s definition of well-being (mentioned earlier),
2) IEEE 1752.1-2021 Standard for Open Mobile Health Data--Representation of Metadata, Sleep, and Physical Activity Measures,
3) ISO/IEEE 11073 medical device communication standard, and
4) IEEE P1157 Medical Data Interchange (MEDIX).
\textbf{IEEE 1752.1-2021 Standard for Open Mobile Health Data--Representation of Metadata, Sleep, and Physical Activity Measures}\footnote{\url{https://standards.ieee.org/ieee/1752.1/6982/}} eases semantic interoperability across mobile health sources and provides meaningful description, exchange, sharing, and use of such mHealth data for consumer health, biomedical research, and clinical care stakeholders.
NIST researchers are collaborating with medical device experts to develop standards for medical device communications to enhance semantic interoperability. XML schema and tool based on the \textbf{ISO/IEEE 11073 Medical Device Communication} standard \citep{garguilo2007moving}.
\textbf{IEEE P1157 Medical Data Interchange (MEDIX)} is a standard for communication of medical information between heterogeneous healthcare information systems such as between a patient care system and selected ancillaries in the medical center setting.
IEEE 1157 Standard for Health Data Interchange is designed with health care professionals.

\textbf{Conclusion: Within IEEE 7010, we found a standard explaining the definition of well-being. Understanding well-being will help as a preventive solution to reduce mental health problems.}

\subsection{Standards: Conclusion}
\label{section:Standard_Concluion}

We did not find  standards for mental health in those SDOs: ETSI SmartM2M, ITU/WHO FG-AI4H, and W3C HCLS CG. ISO 215 Health Informatics must be explored further to dig any documents related to mental health. CEN/ISO EN13606 focuses on EHR semantic interoperability. We are involved in "IoT for Stress Management, Good health \& Well-being, a standard under development within ISO SC41 IoT. ISO SC42 AI joint working group “AI-enabled Health Informatics” comprises experts from ISO/IEC/JTC 1/SC 42, IEEE, and the ITU/WHO AI4Health focus group. NIST has an interest in semantic interoperability. IEEE 7010 focused on well-being which encourages us to investigate well-being as a preventive solution to reduce mental health problems.

\section{LOV4IoT Ontology Catalog for IoT-Based Depression and Mental Health KGs}
\label{section:ontologycatalog}

Ontology catalogs such as BioPortal \citep{noy2009bioportal}, Linked Open Vocabularies (LOV) \citep{vandenbusschelov2015} do not cover sensors (Internet of Things). For this reason,  we built the LOV4IoT ontology catalog (introduced in Section   \ref{section:MentalHealthOntologyCatalog}), with a subset specific to mental health and depression (demonstrated in  Table \ref{table:depressionOntologySOTA}), and emotion ontologies explained in \citep{gyrardACCRA2021} \citep{gyrardEKG2022} or LOV4IoT-Health to collect sensor used, etc.
Mapping to Standards such as ETSI SmartM2M SAREF4EHAW is mentioned in Section \ref{Section:MappingETSISAREF}.
Mapping to Standardized Health KGs/Ontologies/Terminologies such as SNOMED-CT, FMA, RXNORM, MedDRA, LOINC, ChEBI, or well-known knowledge graphs such as DBpedia is described in Section \ref{Section:StandardizedHealthTerminologies}.


\subsection{LOV4IoTMental Health Ontology Catalog and Knowledge Graph}
\label{section:MentalHealthOntologyCatalog}
 We have designed an ontology catalog for depression and mental health, called LOV4IoT Mental Health \footnote{\url{http://lov4iot.appspot.com/?p=lov4iot-depression}} (Table \ref{table:depressionOntologySOTA}).



 \begin {table*}
 \resizebox{\textwidth}{!}{%
\begin{tabular}{ | l | l | l | l | l |}
    
    \hline
    \textbf{Authors} &  \textbf{Year} & \textbf{Project} &  \textbf{Ontology-based} & \textbf{Reasoning} \\
    & & &\textbf{project} & \\

\hline 
Hastings et al. \cite{hastings2012representing} &  2012  & Mental Disease (MD) & \checkmark (online code) & No\\
Ceusters et al. 
Smith et al. &    & Ontology &  & \\ 

\hline 
Amoretti et al. \cite{amoretti2019ontologies} &  2019  & Mental Disorder  & \checkmark (online code) & OWL-DL \\
  &    &  Schizophrenia ontology &  &  \\ 

\hline 
Chang et al. \cite{chang2013depression} &  2015  & \textbf{Depression} ontology & \checkmark (but not shared) & Bayesian networks \\ 
Chang et al. \cite{chang2015mobile}, Taiwan & 2013  & &  & Jena rule (Dysthymia), 46 inference rules\\ 

\hline 
Huang et al. \cite{huang2017constructing} &  2017  & DepressionKG & \checkmark (datasets) & No \\ 

\hline 
Hadzic et al \cite{hadzic2008towards} &  2008  & \textbf{Mental Health} Ontology & \checkmark (but not shared) & No \\ 
Perth, Australia &    & disorder, factors, and treatments &  &  \\ 

\hline 
Jung et al. \cite{jung2017ontology} \cite{jung2015development}, Korea  &  2017-2015  & \textbf{Depression} Ontology, adolescent population, Twitter analysis & \checkmark (but not shared) & No \\ 
 



 	\hline
\end{tabular}}
\caption{Ontology-based depression and mental health projects}
\label{table:depressionOntologySOTA}
\end{table*}

\textbf{GENA (Graph of mEntal-health and Nutrition Association) \citep{dang2023gena}}\footnote{https://github.com/ddlinh/gena-db} encodes relationships between nutrition and mental health. 
GENA describes food, biochemicals, and mental illnesses extracting knowledge from PubMed.
GENA consists of 43,367 relationships with concepts such as nutrition, biochemical, mental health, chemical, and disease. 
GENA used ontologies such as Human Disease Ontology (DOID), Chemical Entities of Biological Interest Ontology (CHEBI), Foundational Model of Anatomy (FMA), Disorders cluster (APADISORDERS), Autism Spectrum Disorder Phenotype Ontology (ASDTTO), The FoodOn Food Ontology (FOODON), MFO Mental Disease Ontology (MFOMD), Protein Ontology (PR), and Symptom Ontology (SYMP).
As an example, CHEBI, FMA, is used as explained in Section \ref{Section:StandardizedHealthTerminologies}.


\textbf{DSM-V (Diagnostic and Statistical Manual of Mental Disorders)}\footnote{\url{https://www.psychiatry.org/psychiatrists/practice/dsm}} references more than 70 mental disorders that complement the International Classification of Diseases (ICD). DSM-V helps clinicians and researchers define and classify mental disorders, which can improve diagnoses, treatment, and research. DSM-V provides a checklist form of symptoms for better diagnosis.
As an example, DSM-V is used as explained in Section \ref{section:OtherProjects}.


\textbf{Mental Health Ontology \citep{hadzic2008towards}} comprises three sub-ontologies: 1) disorder/illness types, (2) factors, and (3) treatments. 
Disorders types are anxiety disorder, eating disorder,  childhood disorder, cognitive disorder, mood disorder.
Factors can be:
1) Physical (e.g. vitamin B deficiency, health injury, liver disease),
2) Environmental (E physical, social, financial),
3) Personal (belief, emotion, response).
The need to understand emotions like stress, anger, bitterness, guilt, joy, happiness, peace, and fear, since they directly affect mental health is highlighted by \cite{hadzic2008towards}. 
\textbf{Mental Functioning (MF) Ontology and Mental Disease (MD) Ontology\footnote{\url{https://obofoundry.org/ontology/mfomd.html}} \citep{hastings2012representing}} describe human mental functioning and disease, including mental processes such as cognitive processes and qualities such as intelligence.
MF Ontology is based on Basic Formal Ontology (BFO).
MD ontology covers concepts such as disease, diagnosis, disorder, and addiction. Mental Functioning Ontology and Mental Disease Ontology could be used to map answers from clinical interview questionnaires about mood, psychotic disorders, and related spectrum conditions. Hastings et al. also designed emotion ontology. 
\textbf{Mapping} to \url{https://obofoundry.org/ontology/mfomd.html} \citep{hastings2012representing} is not simple since there are no labels or comments within the ontology code but concepts IRI such as MFOMD\_0000040. Fortunately, the URI are deferenceable which means that if we copy and paste \url{https://ontobee.org/ontology/MFOMD?iri=http://purl.obolibrary.org/obo/MFOMD_0000040} we can get additional information such as the definition: "A diagnosis asserting the presence of an instance of a mental disease in a given organism."
 \textbf{Depression KG \citep{huang2017constructing}} is a disease-centric KG applied to Major Depressive Disorder, which addresses several challenges: (1) Heterogeneity of datasets, (2) text processing, (3) incompleteness, inconsistency, and incorrectness of datasets, and (4) expressive, representation of medical knowledge. Depression KG utilizes rule-based reasoning over the KG, which helps psychiatric doctors without KG expertise.
\textbf{MDepressionKG \citep{fu2021mdepressionkg}} integrates the human microbial metabolism network, human diseases, microbes and other ontologies.
\textbf{Ontology for College Student Mental Health Service (CSMH) \citep{zhang2020understanding}} describes appointments, mental disorders, self-help resources, information for parents, local referral sources, and substance abuse prevention. Some of the information is extracted from two CSMH websites.
\textbf{Ontology for mental disorders - Schizophrenia Spectrum Ontology \citep{amoretti2019ontologies}}, is compliant with DSM-5 descriptions of mental disorders, with a specific focus on Schizophrenia. It comprises 58 classes (Mental\_Disorder, Patient, and Symptom), 5 properties, and 191 axioms. Classes of the Schizophrenia Spectrum category and the associated symptoms are defined.
Future work is planned to address borderline personality disorder or major depression.
\textbf{Ontology for managing mental healthcare network in Brazil \citep{yamada2018proposal}}, based on BFO, used for integration and interoperability between databases and to design a Semantic Web-based Decision Support System (DSS) for a regional mental healthcare network in Brazil for clinical and administrative processes. The challenges of dealing with standardization and low-quality data are highlighted. The competency question is whether a manager wants to know how many people had schizophrenia in the city of São Paulo in 2017 without looking at various systems of hospitals in the city.
\subsection{Mapping to Standards: ETSI SmartM2M SAREF4EHAW}
\label{Section:MappingETSISAREF}
The mapping to the ETSI SmartM2M SAREF4EHAW ontology is already explained within the book chapter "SAREF4EHAW-Compliant Knowledge Discovery and Reasoning for IoT-based Preventive Healthcare and Well-Being" \citep{gyrardsarefhealthbookchapter1}.
The mapping focuses on the sensor type used.


\subsection{Mapping to Standardized Health KGs/Ontologies/Terminologies: SNOMED-CT, FMA, RXNORM, MedDRA, LOINC, ChEBI, MESH, GALEN and DBpedia}
\label{Section:StandardizedHealthTerminologies}
The mapping to health knowledge bases is explained within "Interdisciplinary IoT and Emotion Knowledge Graph-Based Recommendation System to Boost Mental Health" \citep{gyrardEKG2022}.
We mapped hormones and neurotransmitters concepts. We searched key ontologies on the Bioportal ontology catalog, to be mapped with the Emotion KG. We found ontologies such as SNOMED-CT, Mapping Foundational Model of Anatomy (FMA), RXNORM, MedDRA, Logical Observation Identifier Names and Code (LOINC), Medical Subject Headings (MESH), GALEN, and Chemical Entities of Biological Interest Ontology (ChEBI). The mappings of hormones and neurotransmitters are summarized in two Tables "Subset of mapping hormones and neurotransmitters to existing knowledge bases to demonstrate the difficulty of reusing only one knowledge base." DBpedia is also used due to its popularity, and links emotion-related concepts to existing emotion ontologies when available online. Most of the emotion ontologies cannot be found on BioPortal; only Hastings's ontology \citep{hastings2011emotion} is referenced on BioPortal.


\section{Ontology-Based Mental Health Recommender System and Project Use Cases: ACCRA, etc. }
\label{section:UseCases}

Project use cases are explained in this Section \ref{section:UseCases}. Social robots to support active and healthy aging (ACCRA European-Japan Project) in Section \ref{Section:ACCRA}, Large Language Models (LLMs) for Mental Health in Section \ref{section:LLMProject}, and other projects on Mental Health such as Depression and Suicide in Section \ref{section:OtherProjects}.
\subsection{ACCRA European-Japan Project: Social robots to support active and healthy aging}
\label{Section:ACCRA}
To design emotional-based robotic applications, how can Internet of Robotic Things (IoRT) technology and co-creation methodologies be used? The ACCRA (Agile Co-Creation of Robots for Ageing) EU project\footnote{\url{https://www.accra-project.org/en/sample-page/}} \citep{gyrardACCRA2021}, coordinated by Trialog, develops advanced social robots to support active and healthy aging, co-created by various stakeholders such as aging people and physicians. Three robots, Buddy, ASTRO, and RoboHon, are used for daily life, mobility, and conversation. The three robots understand and convey emotions in real-time using the Internet of Things and Artificial Intelligence technologies (e.g., knowledge-based reasoning).
The ACCRA project explains that social companion robots assist elderly people in staying independent at home and decrease their social isolation. A challenge was to design applications usable by elderly people using co-creation methodologies involving multiple stakeholders and a multidisciplinary research team (e.g., elderly people, medical professionals, and computer scientists such as roboticists or IoT engineers).

\subsection{Large Language Models (LLMs) for Mental Health}
\label{section:LLMProject}

Large Language Models (LLMs) have become increasingly popular due to their vast statistical knowledge, allowing them to produce fluent English sentences and exhibit human-like performance across tasks such as question-answering, summarization, and recommendations. The debut of ChatGPT on November 30, 2022, garnered considerable attention alongside similar autoregressive LLMs like Google BARD, Google GEMINI, and Anthropic Claude \citep{minaee2024large}. Despite their impressive performance, these LLMs have been criticized for providing confidently asserted yet factually inaccurate information, referred to as ``hallucination'' \citep{rawte2023survey}. This poses a significant challenge to their reliability and trustworthiness. Moreover, LLMs sometimes yield inconsistent answers, eroding trust in their outputs. Attention explanations (generated by models) do not closely align with the ground truth explanations provided by human experts \citep{mohammadi2024welldunn}. When LLMs offer irrelevant explanations, it exacerbates trust issues, casting doubt on their reliability as tools \citep{zhang2023siren}.  Consequently, their application in healthcare, particularly mental health, has been hindered. Although there are mental health-specific LLMs, their performance lacks consistency, reliability, explainability, and trust assessment \citep{gaur2024building}.

Efforts to enhance these LLMs for better utility in healthcare have led to recent research focusing on instruction-tuned and retrieval-augmented (e.g. \citet{tilwani2024reasons}) LLMs \citep{lewis2020retrieval}. Instruction tuning involves incorporating an additional feature called ``instruction'' into the dataset, which can be a guideline, protocol, or rule for the LLM to follow \citep{zhang2023instruction, sheth2022process}. However, such training methods have not proven very effective in sensitive domains. Recent work has shown that rules learned by LLMs after instruction tuning do not match the model's performance level, raising concerns about this training approach, particularly in areas akin to moral question answering, such as mental health. An empirical study by \cite{gupta2022learning} demonstrates that LLMs struggle to complete clinical questionnaires for depression and anxiety . Subsequently, \cite{roy2023process} proposed architecture changes, particularly in autoregressive language models, suggesting that a tree-based learning prediction layer could yield safer outcomes in mental health contexts . Further exploration at the intersection of instruction-tuned LLMs and mental health is warranted. Examples of publicly available mental health LLMs include ChatCBPTSD, Diagnosis of Thought Prompting \citep{chen2023empowering}, Mental-LLM (Alpaca/FLAN-T5 based), MentaLLaMA (LLaMA-2 based; \citep{yang2023mentalllama}), ChatCounselo, ExTES-LLaMA (both LLaMA based), and BBMHR (BlenderBot-BST based), while some, like MindShift, Psy-LLM, and LLM-Counselors (all GPT-3.5 based), remain unavailable to the public \citep{hua2024large}.

Retrieval-augmented LLMs represent another category, where a generator model is paired with a knowledge retriever capable of accessing documents from a vectorized database. These LLMs draw context from the retriever, offering reliability and domain-specific explainability crucial in domains like mental health. \cite{gaur2022iseeq} demonstrated the extension of these LLMs to knowledge graphs . However, experimentation has primarily been limited to open-domain knowledge-intensive language understanding tasks, leaving its utility in mental health as an ongoing research question \citep{sarkar2023review}. An example of retrieval-augmented LLM is shown in openCHA \citep{yang2024chatdiet}. \citet{abbasian2023conversational} developed openCHA, a framework that empowers health agents to enhance the processing of healthcare inquiries by efficiently analyzing input queries, integrating essential information, and offering personalized, context-aware responses.The framework's effectiveness in managing complex healthcare tasks through various expert-provided demonstrations. For example, \citet{yang2024chatdiet} leveraged openCHA to create a personalized nutrition-oriented food recommendation chatbot, enabled by user’s longitudinal data on diabetes, American Diabetes Association dietary guidelines, the Nutritionix information, personal causal models, and population models. On an evaluation includes 100 diabetes-related questions on daily
meal choices and the potential risks associated with the diet, openCHA demonstrated superior performance compared to state-of-the-art GPT 4.

\subsection{Other Projects on Mental Health}
\label{section:OtherProjects}

Mental health professionals (MHPs) are overwhelmed by the rising prevalence of declining mental health (MH), depression, and suicide risk. Traditionally, they rely on time-consuming clinical questionnaires (e.g., DSM-5 assessment measures, strengths and difficulties questionnaires) and long patient interviews\footnote{\url{https://wiki.aiisc.ai/index.php?title=Mental_Health_Projects}}. However, the growing demand for MH services and the shortage of MHPs motivate automated methods for early screening. However, it's important to note that receiving high-quality assistance is not guaranteed\footnote{\url{https://wiki.aiisc.ai/index.php?title=Modeling_Social_Behavior_for_Healthcare_Utilization_in_Depression}}. We require online screening assistance as an improvement over online consultations (e.g., BetterHelp), aiming for higher-quality support. AI has emerged as a promising tool for analyzing text data from various sources, including Electronic Health Records (EHR) and social media posts \citep{zhang2022natural, joyce2023explainable, thiruvalluru2021comparing}.

Research in AI and Mental health can be broadly categorized into two categories: (a) Statistical data-driven machine learning in the realm of mental health, as exemplified by the contributions of \cite{de2013predicting}, \cite{de2016discovering}, \cite{saha2019prevalence}, \cite{chancellor2020methods}, \cite{shing2018expert} and \cite{gkotsis2017characterisation}, represents a significant advancement. However, while these studies showcase the effectiveness of AI in this domain, they are limited in their ability to instill trustworthiness, primarily due to a deficiency in explainability and interoperability, which could be greatly enhanced by incorporating domain-specific expertise.
(b) Knowledge-driven Machine Learning for Mental Health: This comprises work involving clinical questionnaires for question answering, summarization, longitudinal assessment \citep{alambo2019question, gupta2022learning, gaur2021characterization, manas2021knowledge}, a diagnostic statistical manual for mental health disorders for identifying mental health disorders \citep{gaur2018let}, and detecting and assessing the severity of substance use disorder using domain-specific drug abuse ontology \citep{kursuncu2018s, lokala2022computational, lokala2022drug}. Machine-readable mental health knowledge has resulted in explainable classification and interpretable design of black-box language models and conversational agents \citep{gaur2021semantics, dalal2024cross, roy2023alleviate}. Further, datasets designed using such a knowledge are capable of examining grounding, instructability, and alignment of domain-specific language models (e.g., ClinicalBERT \citep{alsentzer2019publicly}, PsychBERT \citep{vajre2021psychbert}, MentalBERT \citep{ji2021mentalbert}) to the intent, needs, and requirements of MHPs \citep{sheth2021knowledge, gaur2019knowledge}.  

\section{Conclusion and Future Work}
\label{section:ConclusionFutureWork}

A \textbf{Mental Health KG} (ontology and dataset) acquires knowledge from ontology-based projects classified within the LOV4IoT ontology catalog (Depression, Mental Health, and Emotion).
LOV4IoT supports researchers with 1) the Systematic Literature Survey, which is a time-consuming task and requires an eagerness to learn and investigate existing projects, 2) FAIR principles to encourage researchers to share their reproducible experiments by publishing online ontologies, datasets, rules, etc.  

\textbf{Short-term challenges:} LOV4IoT is relevant for the IoT community. The results are encouraging to update the dataset with additional domains and ontologies.
LOV4IoT leads to the AIOTI (The Alliance for the Internet of Things Innovation) IoT ontology landscape survey form\footnote{\url{https://ec.europa.eu/eusurvey/runner/OntologyLandscapeTemplate}} and analysis result\footnote{\url{https://bit.ly/3fRpQUU}}, executed by the Standard WG - Semantic Interoperability Expert Group. It aims to help industrial practitioners and non-experts answer those questions: Which ontologies are relevant in a certain domain? Where to find them? How to choose the most suitable? Who is maintaining and taking care of their evolution? There is also the AIOTI Health WG white paper publications on health data space "IoT/Edge Computing and Health Data and Data Spaces"\footnote{\url{https://aioti.eu/aioti-white-paper-iot-edge-computing-and-health-data-and-data-spaces/}}, "AI for better health"\footnote{\url{https://aioti.eu/aioti-wg-health-white-paper-on-ai-for-better-health/}}, and "IoT Improving Healthy Urban Living"\footnote{\url{https://aioti.eu/wp-content/uploads/2022/09/IoT-and-Healthy-Urban-Living-Final.pdf}}.

\textbf{Mid-term challenges:}
Automatic knowledge extraction from ontologies and scientific publications describing the ontology purpose is challenging, as highlighted in our AI4EU Knowledge Extraction for the Web of Things (KE4WoT) Challenge. The challenge encourages the reuse of the expertise designed by domain experts and makes the domain knowledge usable, interoperable, and integrated by machines. We released the set of ontologies, as dumps, web services, and tutorials, and made them available.

\textbf{Long-term challenges:} To improve the veracity and the evaluation of the KG integrated with a reasoning engine, involving domain experts such as psychologists, neuroscientists, etc. would enhance the KG, by proving more of the facts.
The knowledge-based reasoning engine can be extended by considering additional research fields such as psychophysiology, psychobiology, etc. An emphasis on the emotional aspect can be done (e.g., fear, pessimism, sadness) since it impacts mental health. 




\section*{Acknowledgements and Funding}
We want to acknowledge the Kno.e.sis research team (lead by Professor Amit Sheth) from Wright State University, Ohio, USA for fruitful discussions about related topics such as "Mental Health/Depression/Suicide", and "Semantic, Cognitive, and Perceptual Computing" and with cognitive psychologists such as Professor Valerie Shalin during Dr. Gyrard's post-doc in 2018-2019. 








This work has partially received funding from the European Union's Horizon 2020 research and innovation program under project grant agreement StandICT.eu 2026 No. 101091933 (open call). We would like to thank the project partners for their valuable comments. The opinions expressed are those of the authors and do not reflect those of the sponsors.








\bibliographystyle{apacite}
\bibliography{MentalHealth_TaylorFrancisTemplate}










\end{document}